\documentclass[letterpaper]{article} 
\usepackage[preprint]{aaai2027}  
\usepackage[hyphens]{url}  
\usepackage{graphicx} 
\usepackage{natbib}  
\usepackage{caption} 
\usepackage{algorithm}
\usepackage{algorithmic}
\usepackage{multirow}

\usepackage{newfloat}
\usepackage{listings}
\DeclareCaptionStyle{ruled}{labelfont=normalfont,labelsep=colon,strut=off} 
\floatstyle{ruled}
\newfloat{listing}{tb}{lst}{}
\floatname{listing}{Listing}

\usepackage{booktabs}
\usepackage{colortbl}

\title{4D-WAM: 4D Consistent World Modeling for Autonomous Driving}
\author{
\normalfont
\mbox{Jiacheng Fu\textsuperscript{\rm 1}\equalcontrib},
\mbox{Yibo Yuan\textsuperscript{\rm 2}\equalcontrib},
\mbox{Meng Tian\textsuperscript{\rm 3}},
\mbox{Yue Li\textsuperscript{\rm 3}},
\mbox{Jiangtong Zhu\textsuperscript{\rm 3}},
\mbox{Jianhua Han\textsuperscript{\rm 3}},
\mbox{Yueyi Zhang\textsuperscript{\rm 4}},
\mbox{Jianwu Fang\textsuperscript{\rm 2}},
\mbox{Jianru Xue\textsuperscript{\rm 2}},
\mbox{Hang Xu\textsuperscript{\rm 3}},
\mbox{Zhiwei Xiong\textsuperscript{\rm 1}\corresponding}
}

\affiliations{
\mbox{\textsuperscript{\rm 1}University of Science and Technology of China}
\quad
\mbox{\textsuperscript{\rm 2}Xi'an Jiaotong University}
\quad

\mbox{\textsuperscript{\rm 3}Yinwang Intelligent Technology Co., Ltd.}
\quad
\mbox{\textsuperscript{\rm 4}Midea Group}
\\[0.35em]
\mbox{jc\_fu@mail.ustc.edu.cn}
\quad
\mbox{zwxiong@ustc.edu.cn}
}

\begin{document}

\maketitle

\begin{abstract}
Emerging World-Action Models (WAMs) have demonstrated promising performance in autonomous driving by jointly modeling future driving scene evolution and trajectory planning. However, existing WAMs are typically trained with video data, which is only 2D projections of the underlying 4D driving scene. Consequently, WAMs fail to understand and capture the structure of 4D scenes and thus generate visually plausible yet 4D inconsistent future predictions that mislead downstream planning. To alleviate this issue, we present 4D-WAM, a model that leverages geometric foundation models for training-time supervision to enable 4D consistent world modeling. Specifically, we feed WAM-predicted future frames into a geometric foundation model, and use 4D-aware responses to define a 4D consistency loss. This loss encourages the model to understand, represent, and predict physically consistent 4D scenes during training, without additional inference cost. Moreover, we identify an early-decision phenomenon in WAMs and propose a decision-oriented timestep sampling strategy that emphasizes supervision at early, high-noise stages, where driving decisions are primarily formed. By propagating 4D supervision to this critical decision-formation phase, the proposed strategy further improves trajectory planning. Extensive experiments demonstrate that 4D-WAM effectively models 4D consistent scene evolution and achieves state-of-the-art performance on challenging NAVSIM-v1 and NAVSIM-v2 benchmarks.
\end{abstract}


\section{Introduction}

Autonomous driving (AD) systems aim to perceive complex traffic scenes, reason about surrounding participants, and generate safe, comfortable, and rule-compliant behaviors. Vision-language-action (VLA) models have been widely explored in recent years as a paradigm for end-to-end AD, leveraging large-scale multimodal pretraining to unify visual understanding, language-level reasoning, and trajectory or control generation within a single policy \cite{drivegpt4,drivevlm,lmdrive,opendrivevla}.

However, VLAs face two fundamental limitations in trajectory planning. First, although language excels at semantic abstraction and high-level reasoning, it is not naturally suited to the continuous, fine-grained spatiotemporal reasoning demanded by driving. Second, linguistic reasoning in existing VLAs remains only loosely coupled with trajectory prediction, limiting its effectiveness \cite{drive-R1}. These two problems are difficult to overcome by merely scaling up action-labeled data.

\begin{figure}[!t]
\centering
\includegraphics[width=\columnwidth]{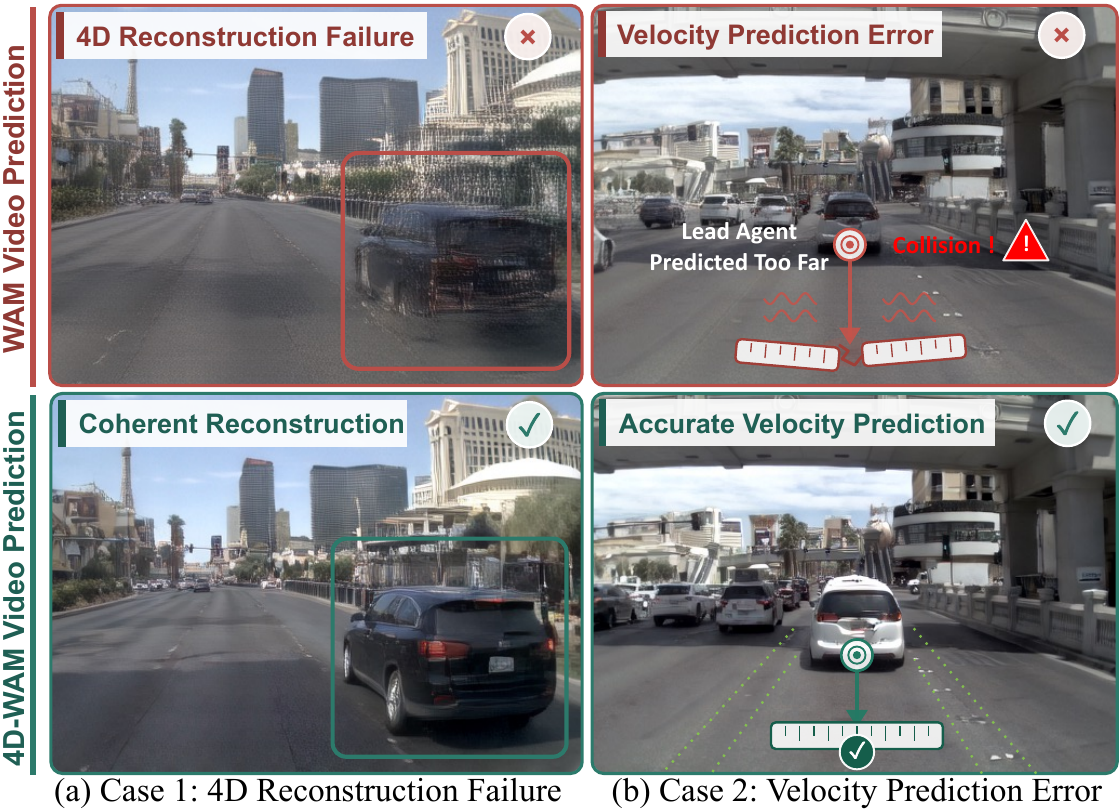}
\caption{Existing WAMs, trained with 2D video supervision, fail to capture the structure of underlying 4D driving scenes. This leads to two
critical failure modes in complex driving scenarios. \textbf{(a) 4D
Reconstruction Failure.} Highly dynamic objects exhibit visual artifacts and
geometric deformation. \textbf{(b) Velocity Prediction Error.} Lacking reliable distance perception, WAMs fail to predict the lead agent's velocity, resulting in
a collision. Through 4D consistency supervision, 4D-WAM effectively addresses both failure modes and achieves 4D consistent world modeling.}
\label{fig1}
\end{figure}

World-Action Models (WAMs) have emerged as a new paradigm to address these limitations. Rather than mapping observations directly to actions, WAMs jointly model and predict future scene evolution and actions. Such joint modeling provides denser spatiotemporal supervision and stronger foresight into scene evolution and surrounding-agent motion, thereby improving planning, particularly in complex driving scenarios \cite{drivevlaw0, mimicvideo, dreamzero, motus}. However, existing WAMs are supervised by videos, where each frame is only a 2D projection of the underlying dynamic 4D driving scene. This raises a fundamental question: 
\textit{ WAMs may produce visually plausible predictions, but do they truly understand the 4D world and ensure that the predicted frames correspond to a physically consistent 4D scene?} 
As demonstrated in \cite{geometry_forcing}, depth-head probing of the intermediate features of world models reveals that these models struggle to understand and capture the underlying structure of a 4D scene. Consequently, their predictions fail to preserve a geometrically coherent scene layout, including accurate depth ordering, spatial relationships, and object-level occlusion patterns. In the context of WAMs, such unreliable predictions can severely mislead downstream trajectory planning. We further reveal this issue in Fig. \ref{fig1}.

This motivates us to propose \textbf{4D-WAM}, a world-action model that realizes 4D consistent world modeling through training-time supervision from geometric foundation models. The key insight is that recent advanced geometric foundation models, such as VGGT-$\Omega$ \cite{vggt_omega}, can faithfully reconstruct 4D scenes from 2D video frames and thus, can act as geometric teachers for 4D consistency evaluation and supervision. Specifically, 4D-WAM feeds the predicted future frames and the ground-truth future frames separately into a frozen geometric foundation model, and uses the resulting 4D-aware responses to define a 4D consistency loss. This loss consists of a feature loss that supervises global scene layout and cross-frame consistency, and a depth loss that further constrains fine-grained geometric relationships. The resulting 4D training signal encourages 4D-WAM to understand, represent, and predict 4D driving scenes, notably without introducing any additional inference cost. Moreover, we uncover an early-decision phenomenon in WAMs: both video and action branches form their driving decisions within the first one or two high-noise denoising steps, while subsequent steps mainly refine visual details and smooth the predicted trajectory. Building on this observation, we propose a Decision-Oriented Timestep Sampling strategy. It identifies a decision point, partitions the denoising process into decision and refinement regions, and allocates more timestep sampling to the decision region. This reallocation allows the proposed 4D supervision to more effectively guide the formation of driving decisions. Extensive experiments on NAVSIM-v1 and NAVSIM-v2 demonstrate that 4D-WAM achieves state-of-the-art planning performance across both standard and challenging driving scenarios.

Our contributions are summarized as follows:
\begin{itemize}
    \item We develop a driving-oriented WAM backbone with an asymmetric attention mask that preserves reliable historical conditioning and enables sufficient video-action interaction.
    
    \item We introduce 4D consistency supervision that aligns predicted futures with geometric features and depth cues from a frozen geometric foundation model.
    
    \item We reveal the early-decision phenomenon in WAMs and propose Decision-Oriented Timestep Sampling to emphasize the decision region during training.
    
    \item 4D-WAM achieves state-of-the-art performance on NAVSIM-v1 \textit{navtest}, NAVSIM-v2 \textit{navtest}, and NAVSIM-v2 \textit{navhard}.
\end{itemize}

\begin{figure*}[!t]
\centering
\includegraphics[width=\textwidth]{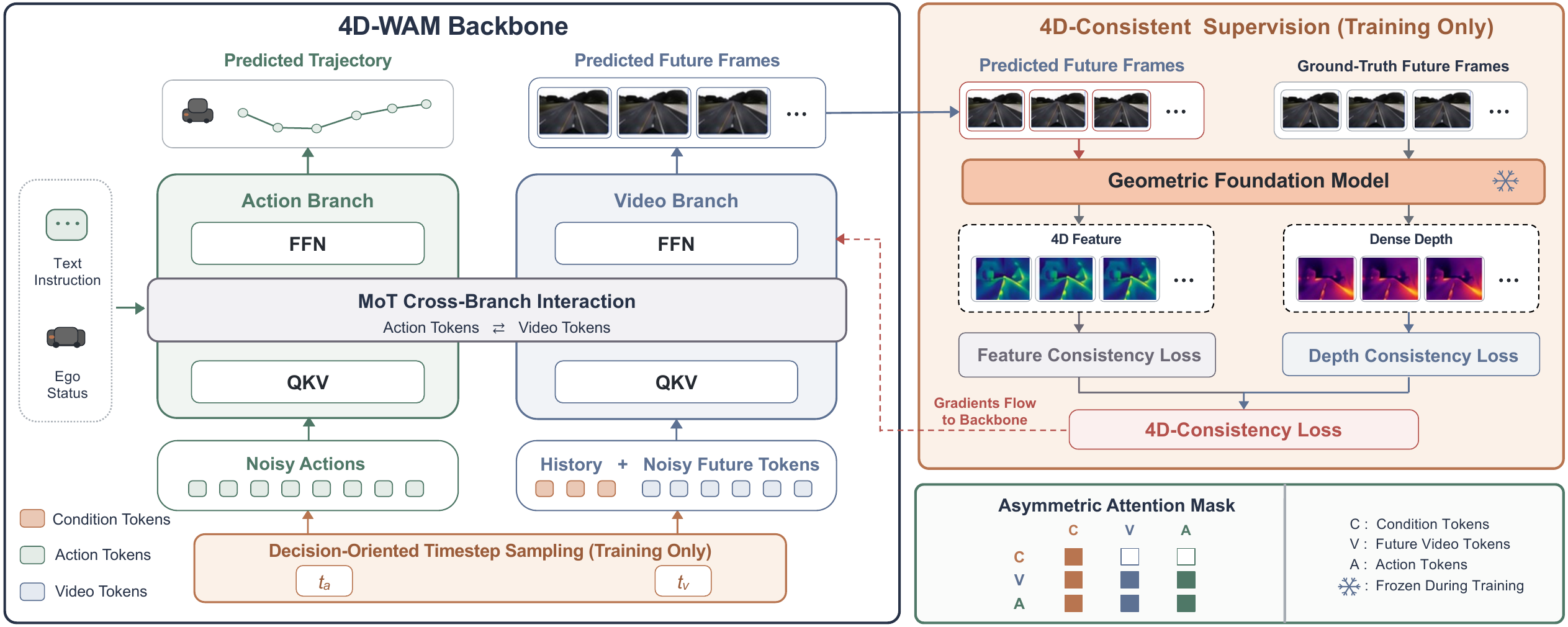}
\caption{
Overview of 4D-WAM.
The MoT-based backbone (left) jointly predicts future video frames
and the ego trajectory, with decision-oriented timestep sampling
applied during training.
A frozen geometric foundation model (top right) extracts 4D features
and dense depth from both predicted and ground-truth future frames,
providing 4D consistency supervision to the backbone through
feature- and depth-consistency losses.
The asymmetric attention mask (bottom right) prevents condition tokens
from attending to future video and action tokens while enabling
bidirectional video--action interaction.
    }
\label{fig2}
\end{figure*}

\section{Related Work}

\subsection{World-Action Models}
Existing WAMs can be broadly grouped into two lines: VLM-based WAMs and video-generation-based WAMs. The first line extends VLA policies with future-scene prediction. VLA-World~\cite{vlaworld} generates future observations from action-derived trajectories and refines planning by reasoning over the imagined future; DriveVLA-W0~\cite{drivevlaw0} jointly learns future-image prediction and action generation, using the former as dense self-supervision to capture driving dynamics and provide richer representations for an efficient action expert. The second line builds WAMs upon video generative models, jointly modeling video and action prediction. PAD~\cite{pad} jointly denoises future images and actions; UWM~\cite{uwm} couples video and action diffusion with decoupled timesteps in a shared transformer; and DreamZero~\cite{dreamzero} scales this paradigm with large video diffusion backbones. These methods demonstrate the advantage of jointly modeling future observations and actions, as dense spatio-temporal prediction can improve world representations and benefit action prediction. However, models in both lines are still supervised by RGB videos or video latents, where each frame is only a 2D projection of the underlying dynamic 4D world; consequently, video-only supervision may yield visually plausible but geometrically or physically inconsistent futures~\cite{geometry_forcing,gem4d}. Our work addresses this limitation by introducing 4D-aware geometric supervision into WAM training.

\subsection{Feed-Forward 3D and 4D Reconstruction}Recent feed-forward reconstruction methods have shifted 3D perception from cascaded geometric pipelines toward learned models that infer scene structure directly from RGB observations. DUSt3R~\cite{dust3r} introduced point-map regression for uncalibrated image pairs, while MASt3R~\cite{mast3r} augmented this representation with dense local descriptors for accurate 3D-grounded matching, and VGGT~\cite{vggt} scaled this paradigm to hundreds of views by jointly predicting cameras, depth, point maps, and point tracks in a single forward pass. Subsequent work extended these models beyond static scenes: DynamicVGGT~\cite{dynamicvggt} explicitly models temporal correspondence and point motion for driving-scene reconstruction. More recently, VGGT-$\Omega$~\cite{vggt_omega} has advanced this line through large-scale supervised training, together with a register-based architecture designed for efficient reconstruction. These advances suggest that geometric foundation models encode transferable spatial and temporal structure from RGB observations. In 4D-WAM, we innovatively use a frozen VGGT-$\Omega$ as a training-time geometric teacher to provide 4D-aware supervision signal.

\section{Method}

\subsection{4D-WAM Architecture} 

The overall architecture of 4D-WAM is shown in Fig.~\ref{fig2}. Given historical front-view frames
$\mathbf{v}_{-T_h:0}
=\{\mathbf{v}_{-T_h},\ldots,\mathbf{v}_0\}$,
the current ego state $\mathbf{e}_0$, and a textual driving
instruction $q$, 4D-WAM jointly models future scene evolution
and ego planning through the conditional distribution:
\begin{equation}
p_{\phi}\!\left(
\mathbf{v}_{1:T_f},
\mathbf{a}_{1:T_p}
\mid
\mathbf{v}_{-T_h:0},
\mathbf{e}_0,
q
\right),
\end{equation}
where $\phi$ denotes the learnable model parameters,
$\mathbf{v}_{1:T_f}
=\{\mathbf{v}_t\}_{t=1}^{T_f}$
represents the future video sequence, and
$\mathbf{a}_{1:T_p}
=\{\mathbf{a}_t\}_{t=1}^{T_p}$
denotes the planned ego trajectory. The historical frames are encoded by the video VAE into clean history latents that serve as a condition prefix. These latents are concatenated with noisy future-video latents and fed to the
video branch. In parallel, a noisy action trajectory is linearly
projected into action tokens and fed to the action branch. The
textual instruction is encoded by a pretrained T5 text encoder,
while the ego state is linearly projected and appended to the
text tokens. 

To preserve modality-specific representations while enabling
joint reasoning, 4D-WAM adopts a Mixture-of-Transformers (MoT)
design with separate video and action experts. The video expert
is initialized from a pretrained Wan2.2 video DiT \cite{wan2.2}, while the action expert employs a structurally aligned action DiT. At the
$\ell$-th layer, the video expert computes query, key, and value
representations for the concatenated history and future-video
tokens, denoted by
$(\mathbf{Q}^{\ell}_{v},\mathbf{K}^{\ell}_{v},
\mathbf{V}^{\ell}_{v})$. The action expert computes the
corresponding representations for the action tokens, denoted by
$(\mathbf{Q}^{\ell}_{a},\mathbf{K}^{\ell}_{a},
\mathbf{V}^{\ell}_{a})$. We concatenate the representations
along the token dimension,
\begin{equation}
\mathbf{Q}^{\ell}
=[\mathbf{Q}^{\ell}_{v};\mathbf{Q}^{\ell}_{a}],
\quad
\mathbf{K}^{\ell}
=[\mathbf{K}^{\ell}_{v};\mathbf{K}^{\ell}_{a}],
\quad
\mathbf{V}^{\ell}
=[\mathbf{V}^{\ell}_{v};\mathbf{V}^{\ell}_{a}],
\end{equation}
and perform an asymmetric masked mixed attention as
\begin{equation}
\widetilde{\mathbf{X}}^{\ell}
=
\mathrm{Attn}\!\left(
\mathbf{Q}^{\ell},
\mathbf{K}^{\ell},
\mathbf{V}^{\ell};
\mathbf{M}
\right).
\end{equation}
The resulting attention output is partitioned according to the original token boundaries and routed back to the respective experts for subsequent modality-specific processing. 

We carefully design the asymmetric attention mask $\mathbf{M}$ to preserve reliable conditioning while enabling sufficient interaction between the two branches. Visual condition tokens attend only to themselves, maintaining the observed history as a stable anchor and preventing the interference of noisy tokens. In contrast, video tokens and action tokens can attend to all three token groups, allowing both branches to be fully conditioned on the visual history while exchanging complementary information throughout joint prediction. 

\begin{figure*}[t]
\centering
\includegraphics[width=\textwidth]{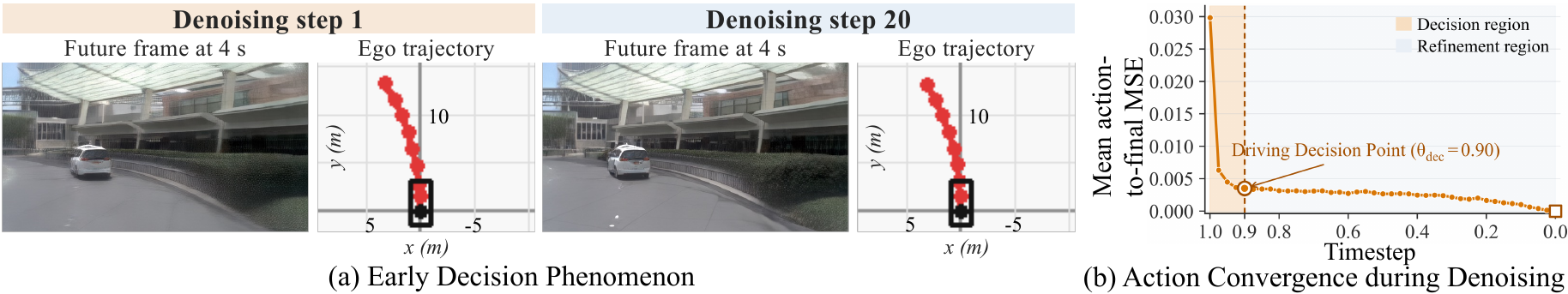}
\caption{\textbf{Early decision phenomenon in WAMs.}
(a) Both video and action branches finalize driving decisions in an
extremely high-noise region.
(b) The action-to-final MSE rapidly decreases and plateaus at
$\theta=0.90$, marking the driving decision point that separates the decision
and refinement regions.}
\label{early_decision_viz}
\end{figure*}


\subsection{4D Consistency Supervision}

As discussed, existing WAMs often fail to capture and represent the underlying structure of 4D scenes. Consequently, physically inconsistent video predictions will mislead trajectory prediction. To address this limitation, we propose to leverage geometric foundation models to provide 4D consistency supervision during training.

Geometric foundation models operate on clean RGB frames, whereas directly decoding noisy future-video latents is infeasible since the video VAE decoder is highly sensitive to latent perturbations. We therefore first recover a clean-sample estimate before VAE decoding. Let $\sigma_v\in[0,1]$ denote the normalized video noise level, where $\sigma_v=0$ corresponds to clean data and $\sigma_v=1$ to pure noise. Given the noisy future-video latent $\mathbf{z}_{\sigma_v}^{v}$, the 
video branch predicts its velocity $\hat{\mathbf{u}}_{\phi}^{v}$. 
The clean-sample latent estimate is then recovered as
\begin{equation}
    \hat{\mathbf{z}}_{0}^{v}
    =
    \mathbf{z}_{\sigma_v}^{v}
    -
    \sigma_v\hat{\mathbf{u}}_{\phi}^{v}.
\end{equation}
Decoding this estimate with the video VAE decoder produces the predicted future video:
\begin{equation}
    \hat{\mathbf{v}}_{1:T_f}
    =
    D_{\mathrm{VAE}}\!\left(\hat{\mathbf{z}}_{0}^{v}\right).
\end{equation}

Next, we separately feed the predicted future video 
$\hat{\mathbf{v}}_{1:T_f}$ and its ground-truth counterpart 
$\mathbf{v}_{1:T_f}^{\mathrm{gt}}$ into the same frozen 
VGGT-$\Omega$~\cite{vggt_omega}, a state-of-the-art 4D geometric 
foundation model denoted by $\mathcal{G}$. For each input, we retain 
its multi-level geometric features and estimated depth maps:
\begin{equation}
\begin{array}{ll}
\big(\{\hat{\mathbf{F}}^{\ell}\}_{\ell\in\mathcal{S}},
\hat{\mathbf{D}}\big)
&=\mathcal{G}(\hat{\mathbf{v}}_{1:T_f}),\\
\big(\{\mathbf{F}_{\mathrm{gt}}^{\ell}\}_{\ell\in\mathcal{S}},
\mathbf{D}_{\mathrm{gt}}\big)
&=\mathcal{G}(\mathbf{v}_{1:T_f}^{\mathrm{gt}}).
\end{array}
\end{equation}

Using the ground-truth responses as supervision targets, we define 
the feature consistency loss as
\begin{equation}
\mathcal{L}_{\mathrm{feat}}
=
\frac{1}{|\mathcal{S}|}
\sum_{\ell\in\mathcal{S}}
d_{\mathrm{cos}}
\left(
\hat{\mathbf{F}}^{\ell},
\mathbf{F}_{\mathrm{gt}}^{\ell}
\right).
\end{equation}

We further define a depth consistency loss over reliable future pixels.
Let $\mathcal{Q}$ denote the pixel set selected based on validity checks
and the confidence estimates produced by VGGT-$\Omega$. For each
$q\in\mathcal{Q}$, the predicted and ground-truth log-depths are given by
$\hat d_q=\log(\hat{\mathbf{D}}_q+\epsilon)$ and
$d_{\mathrm{gt},q}=\log(\mathbf{D}_{\mathrm{gt},q}+\epsilon)$,
respectively. We minimize their discrepancy over the selected pixels:
\begin{equation}
\mathcal{L}_{\mathrm{depth}}
=
\frac{1}{|\mathcal{Q}|}
\sum_{q\in\mathcal{Q}}
\mathrm{SmoothL1}_{\beta}
\left(
\hat d_q-d_{\mathrm{gt},q}
\right).
\end{equation}

The feature- and depth-consistency losses provide complementary supervision for capturing the underlying 4D scene structure. Feature consistency captures global scene layout, inter-frame dynamics, and object correspondences across frames. Depth consistency complements it by constraining dense pixel-level geometry, including precise relative depth ordering, object boundaries, and fine-grained spatial structure. We combine the feature- and depth-consistency losses and propose the 4D consistency loss as
\begin{equation}
\mathcal{L}_{\mathrm{4D}}
=
\lambda_{\mathrm{feat}}\mathcal{L}_{\mathrm{feat}}
+
\lambda_{\mathrm{depth}}\mathcal{L}_{\mathrm{depth}}.
\end{equation}

\begin{table*}[!t]
\centering
\small
\setlength{\tabcolsep}{1.05mm}
\renewcommand{\arraystretch}{1.06}

\resizebox{\linewidth}{!}{%
\begin{tabular}{@{}lccccccccccc@{}}
\toprule
\textbf{Method}
& \textbf{Sensors}
& \textbf{NC} $\uparrow$
& \textbf{DAC} $\uparrow$
& \textbf{DDC} $\uparrow$
& \textbf{TLC} $\uparrow$
& \textbf{EP} $\uparrow$
& \textbf{TTC} $\uparrow$
& \textbf{LK} $\uparrow$
& \textbf{HC} $\uparrow$
& \textbf{EC} $\uparrow$
& \textbf{EPDMS} $\uparrow$ \\
\midrule

\rowcolor[gray]{0.94}
\multicolumn{12}{c}{\textit{Reference}} \\

Human Agent
& --
& 100.0
& 100.0
& 99.8
& 100.0
& 87.4
& 100.0
& 100.0
& 98.1
& 90.1
& 94.5 \\

\midrule
\rowcolor[gray]{0.94}
\multicolumn{12}{c}{\textit{Vision-Based End-to-End Methods}} \\

TransFuser~\cite{transfuser}
& $3\mathrm{C}+\mathrm{L}$
& 96.9
& 89.9
& 97.8
& 99.7
& 87.1
& 95.4
& 92.7
& \underline{98.3}
& 87.2
& 76.7 \\

DriveSuprim~\cite{drivesuprim}
& $3\mathrm{C}$
& 97.5
& 96.5
& 99.4
& 99.6
& 88.4
& 96.6
& 95.5
& \underline{98.3}
& 77.0
& 83.1 \\

DiffusionDrive~\cite{diffusiondrive}
& $3\mathrm{C}+\mathrm{L}$
& 98.2
& 95.9
& 99.4
& \underline{99.8}
& 87.5
& 97.3
& 96.8
& \underline{98.3}
& \underline{87.7}
& 84.5 \\

\midrule
\rowcolor[gray]{0.94}
\multicolumn{12}{c}{\textit{Vision-Language-Action Methods}} \\

ReCogDrive$^{\mathrm{RL}}$~\cite{recogdrive}
& $1\mathrm{C}$
& 98.3
& 95.2
& 99.5
& \underline{99.8}
& 87.1
& 97.5
& 96.6
& \underline{98.3}
& 86.5
& 83.6 \\

SGDrive~\cite{sgdrive}
& $1\mathrm{C}$
& 98.6
& 94.3
& 99.5
& \textbf{99.9}
& 86.0
& 97.9
& 96.1
& \underline{98.3}
& 85.9
& 86.2 \\

DriveFine~\cite{drivefine}
& $1\mathrm{C}$
& \underline{98.7}
& 97.3
& 99.5
& \underline{99.8}
& \underline{88.7}
& 97.8
& \underline{97.7}
& \textbf{98.4}
& 83.8
& \underline{89.7} \\

\midrule
\rowcolor[gray]{0.94}
\multicolumn{12}{c}{\textit{World-Action Models}} \\

Epona~\cite{epona}
& $1\mathrm{C}$
& 97.1
& 95.7
& 99.3
& \underline{99.8}
& 88.6
& 96.3
& 97.0
& 98.0
& 67.8
& 85.1 \\

DriveVLA-W0~\cite{drivevlaw0}
& $1\mathrm{C}$
& 98.5
& \textbf{99.1}
& 98.0
& 99.7
& 86.4
& 98.1
& 93.2
& 97.9
& 58.9
& 86.1 \\

WorldRFT$^{\mathrm{RL}}$~\cite{worldrft}
& $3\mathrm{C}$
& 97.8
& 96.5
& 99.5
& \underline{99.8}
& 88.5
& 97.0
& 97.4
& 98.1
& 69.1
& 86.7 \\

DriveLaW~\cite{drivelaw}
& $1\mathrm{C}$
& \underline{98.7}
& 96.9
& \underline{99.6}
& \underline{99.8}
& 87.5
& \underline{98.3}
& 97.6
& \textbf{98.4}
& 77.4
& 88.6 \\

Metis~\cite{metis}
& $1\mathrm{C}$
& 98.4
& 97.2
& \underline{99.6}
& \underline{99.8}
& 87.8
& 97.7
& \textbf{97.8}
& \textbf{98.4}
& \textbf{88.0}
& 89.5 \\

\addlinespace[0.15em]
\textbf{4D-WAM (Ours)}
& $1\mathrm{C}$
& \textbf{99.1}
& \underline{97.9}
& \textbf{99.7}
& 97.3
& \textbf{98.3}
& \textbf{98.6}
& 86.9
& 97.7
& 87.5
& \textbf{90.6} \\

\bottomrule
\end{tabular}%
}

\caption{Comparison on the \textit{navtest} split of NAVSIM v2.
Bold and underlined values denote the best and second-best results
among learned methods, respectively.
$^{\mathrm{RL}}$ denotes methods with reinforcement-learning fine-tuning.}
\label{tab:navsim_v2_epdms}
\end{table*}

\begin{table}[t]
\centering
\small
\setlength{\tabcolsep}{1mm}
\renewcommand{\arraystretch}{1.06}

\resizebox{\columnwidth}{!}{%
\begin{tabular}{@{}lcccccc@{}}
\toprule
\textbf{Method}
& \textbf{NC} $\uparrow$
& \textbf{DAC} $\uparrow$
& \textbf{TTC} $\uparrow$
& \textbf{C} $\uparrow$
& \textbf{EP} $\uparrow$
& \textbf{PDMS} $\uparrow$ \\
\midrule

\rowcolor[gray]{0.94}
\multicolumn{7}{c}{\textit{Reference}} \\

Human Agent
& 100.0
& 100.0
& 100.0
& 99.9
& 87.5
& 94.8 \\

\midrule
\rowcolor[gray]{0.94}
\multicolumn{7}{c}{\textit{Vision-Based End-to-End Methods}} \\

TransFuser~\cite{transfuser}
& 97.7
& 92.8
& 92.8
& \textbf{100.0}
& 79.2
& 84.0 \\

DiffusionDrive~\cite{diffusiondrive}
& 98.2
& 96.2
& 94.7
& \textbf{100.0}
& 82.2
& 88.1 \\

DriveSuprim~\cite{drivesuprim}
& 97.8
& 97.3
& 93.6
& \textbf{100.0}
& \underline{86.7}
& 89.9 \\

\midrule
\rowcolor[gray]{0.94}
\multicolumn{7}{c}{\textit{Vision-Language-Action Methods}} \\

SGDrive~\cite{sgdrive}
& 98.6
& 95.1
& 95.4
& \textbf{100.0}
& 81.2
& 87.4 \\

DriveFine~\cite{drivefine}
& 98.6
& \underline{97.9}
& 95.2
& \underline{99.9}
& 85.5
& 90.7 \\

ReCogDrive$^{\mathrm{RL}}$~\cite{recogdrive}
& 97.9
& 97.3
& 94.9
& \textbf{100.0}
& \textbf{87.3}
& \underline{90.8} \\

\midrule
\rowcolor[gray]{0.94}
\multicolumn{7}{c}{\textit{World-Action Models}} \\

Epona~\cite{epona}
& 97.9
& 95.1
& 93.8
& \underline{99.9}
& 80.4
& 86.2 \\

WorldRFT$^{\mathrm{RL}}$~\cite{worldrft}
& 97.5
& 96.0
& 94.0
& \textbf{100.0}
& 80.9
& 87.0 \\

DriveLaW~\cite{drivelaw}
& \underline{99.0}
& 97.1
& \textbf{96.7}
& \textbf{100.0}
& 81.3
& 89.1 \\

Metis~\cite{metis}
& 98.3
& 97.1
& 94.7
& \textbf{100.0}
& 83.4
& 89.1 \\

DriveVLA-W0~\cite{drivevlaw0}
& 98.7
& \textbf{99.1}
& 95.3
& 99.3
& 83.3
& 90.2 \\

\addlinespace[0.15em]
\textbf{4D-WAM (Ours)}
& \textbf{99.1}
& \underline{97.9}
& \underline{96.6}
& \textbf{100.0}
& 84.6
& \textbf{90.9} \\

\bottomrule
\end{tabular}%
}

\caption{Comparison on the \textit{navtest} split of NAVSIM v1. $^{\mathrm{RL}}$ denotes methods with reinforcement-learning fine-tuning.}
\label{tab:navsim_v1_pdms}
\end{table}

\begin{table*}[t]
\centering
\small
\setlength{\tabcolsep}{1mm}
\renewcommand{\arraystretch}{1.06}

\resizebox{\linewidth}{!}{%
\begin{tabular}{@{}lccccccccccc@{}}
\toprule
\textbf{Method}
& \textbf{Stage}
& \textbf{NC} $\uparrow$
& \textbf{DAC} $\uparrow$
& \textbf{DDC} $\uparrow$
& \textbf{TLC} $\uparrow$
& \textbf{EP} $\uparrow$
& \textbf{TTC} $\uparrow$
& \textbf{LK} $\uparrow$
& \textbf{HC} $\uparrow$
& \textbf{EC} $\uparrow$
& \textbf{EPDMS} $\uparrow$ \\
\midrule

\rowcolor[gray]{0.94}
\multicolumn{12}{c}{\textit{Vision-Based End-to-End Methods}} \\

\multirow{2}{*}{TransFuser~\cite{transfuser}}
& S1
& 96.2
& 79.5
& 99.1
& 99.5
& 84.1
& 95.1
& 94.2
& 97.5
& 79.1
& \multirow{2}{*}{23.1} \\
& S2
& 77.7
& 70.2
& 84.2
& 98.0
& 85.1
& 75.6
& 45.4
& 95.7
& \textbf{75.9}
& \\

\addlinespace[0.10em]

\multirow{2}{*}{DiffusionDrive~\cite{diffusiondrive}}
& S1
& 96.8
& 86.0
& 98.8
& 99.3
& 84.0
& 95.8
& 96.7
& \underline{97.6}
& \underline{79.6}
& \multirow{2}{*}{27.5} \\
& S2
& 80.1
& \underline{72.8}
& 84.4
& 98.4
& 85.9
& 76.6
& 46.4
& 96.3
& 72.8
& \\

\midrule
\rowcolor[gray]{0.94}
\multicolumn{12}{c}{\textit{Vision-Language-Action Methods}} \\

\multirow{2}{*}{SGDrive~\cite{sgdrive}}
& S1
& 95.8
& 87.6
& 97.8
& \textbf{99.8}
& 84.4
& 94.7
& 92.9
& \textbf{97.8}
& 28.9
& \multirow{2}{*}{25.5} \\
& S2
& 79.4
& 65.4
& 79.1
& \textbf{98.9}
& 88.9
& 75.3
& 42.7
& 96.4
& 29.6
& \\

\addlinespace[0.10em]

\multirow{2}{*}{ReCogDrive$^{\mathrm{RL}}$~\cite{recogdrive}}
& S1
& 96.4
& 78.9
& 98.7
& \textbf{99.8}
& 82.6
& 95.6
& 94.4
& \underline{97.6}
& 74.2
& \multirow{2}{*}{25.7} \\
& S2
& 80.2
& 65.0
& 82.4
& \underline{98.7}
& 85.2
& 76.9
& 43.8
& \underline{96.6}
& 71.8
& \\

\addlinespace[0.10em]

\multirow{2}{*}{DriveFine~\cite{drivefine}}
& S1
& \textbf{97.6}
& \textbf{90.0}
& 99.1
& 99.3
& \underline{84.9}
& \underline{96.7}
& \underline{97.3}
& \underline{97.6}
& 72.0
& \multirow{2}{*}{30.5} \\
& S2
& 82.1
& 71.3
& 84.8
& 98.4
& 88.1
& 74.3
& 47.2
& \textbf{96.8}
& 72.8
& \\

\midrule
\rowcolor[gray]{0.94}
\multicolumn{12}{c}{\textit{World-Action Models}} \\

\multirow{2}{*}{DriveVLA-W0~\cite{drivevlaw0}}
& S1
& 96.8
& 83.3
& 99.0
& \underline{99.6}
& 84.6
& 95.3
& 96.4
& \underline{97.6}
& 78.2
& \multirow{2}{*}{24.4} \\
& S2
& 76.8
& 64.3
& 79.9
& 98.3
& \underline{89.2}
& 75.0
& 46.8
& 95.8
& 53.1
& \\

\addlinespace[0.10em]

\multirow{2}{*}{DriveLaW~\cite{drivelaw}}
& S1
& \underline{97.3}
& 89.1
& \underline{99.2}
& \underline{99.6}
& 84.3
& \textbf{97.1}
& 96.2
& \textbf{97.8}
& 67.6
& \multirow{2}{*}{30.6} \\
& S2
& \underline{82.5}
& 67.6
& 83.5
& 98.1
& 84.8
& \underline{78.5}
& 45.8
& 96.4
& 57.3
& \\

\addlinespace[0.10em]

\multirow{2}{*}{Metis~\cite{metis}}
& S1
& 96.6
& 87.8
& 99.0
& 99.3
& 84.5
& 95.6
& \textbf{97.8}
& \textbf{97.8}
& 77.8
& \multirow{2}{*}{\underline{32.2}} \\
& S2
& 79.6
& \textbf{73.3}
& \underline{84.9}
& 97.8
& 85.8
& 76.6
& \underline{47.7}
& 95.4
& \underline{75.3}
& \\

\addlinespace[0.10em]

\multirow{2}{*}{\textbf{4D-WAM (Ours)}}
& S1
& 92.8
& \underline{89.6}
& \textbf{99.7}
& 99.3
& \textbf{98.6}
& 91.6
& 86.9
& 97.1
& \textbf{81.3}
& \multirow{2}{*}{\textbf{35.9}} \\
& S2
& \textbf{82.6}
& 71.8
& \textbf{86.9}
& 98.2
& \textbf{97.6}
& \textbf{78.8}
& \textbf{49.0}
& 95.8
& 71.6
& \\

\bottomrule
\end{tabular}%
}

\caption{Comparison on the NAVSIM v2 \textit{navhard} benchmark.
S1 and S2 denote the real-observation and synthetic follow-up stages,
respectively. EPDMS is the final two-stage score. $^{\mathrm{RL}}$ denotes methods with reinforcement-learning fine-tuning.}
\label{tab:navsim_v2_navhard}
\end{table*}

\subsection{Decision-Oriented Timestep Sampling}

WAMs generate video and action outputs through iterative denoising, yet it remains unclear when the driving decision is actually formed during this process. To examine this process, we construct a 20-step joint denoising rollout and recover the clean-sample estimates after every step. Fig.~\ref{early_decision_viz}(a) visualizes the future video and trajectory estimates. We surprisingly find that for both branches, the driving decision is determined after only one or two high-noise steps. The subsequent low-noise steps largely preserve the selected maneuver, primarily refining visual details or applying minor local smoothing to the trajectory. We refer to this phenomenon as \emph{early decision in WAMs}.

We further quantify this phenomenon through the action branch, whose predicted trajectory directly reflects the model's driving decision. We select 1,000 scenes from the NAVSIM validation set for this analysis. For each scene $i$, we record the action clean-sample estimate $\hat{\mathbf a}_{i,1:T_p}^{(k)}$ at every step $k$ of the 20-step denoising process. We measure the action-to-final discrepancy using the MSE between each intermediate estimate and the final 20-step prediction:
\begin{equation}
D^a(k)
=
\frac{1}{N T_p}
\sum_{i=1}^{N}
\sum_{t=1}^{T_p}
\left\|
\hat{\mathbf a}_{i,t}^{(k)}
-
\hat{\mathbf a}_{i,t}^{(20)}
\right\|_2^2,
\end{equation}
where $N$ denotes the number of evaluated scenes. As shown in Fig.~\ref{early_decision_viz}(b), the action-to-final MSE
drops sharply during the initial denoising steps and reaches a plateau
at $\theta=0.90$. We therefore identify this point as the driving
decision point, denoted by $\theta_{\mathrm{dec}}$, and divide the
denoising process into a decision region and a refinement region:
\begin{equation}
\mathcal{R}_{\mathrm{dec}}
=
[\theta_{\mathrm{dec}},1],
\qquad
\mathcal{R}_{\mathrm{ref}}
=
[0,\theta_{\mathrm{dec}}).
\end{equation}

Motivated by this observation, we propose \textit{Decision-Oriented
Timestep Sampling}, which biases training-time timestep sampling toward
the decision region. This directs 4D consistency supervision more strongly toward learning how driving decisions are formed in joint video-action prediction, rather than toward late-stage video refinement. Specifically, we sample the normalized noise level via
\begin{equation}
\theta
=
\frac{s u}{1+(s-1)u},
\qquad
u\sim\mathcal{U}(0,1).
\end{equation}
Here, $s=1$ corresponds to uniform sampling, whereas $s>1$ shifts the sampling distribution toward higher noise levels. We calibrate $s$ according to the probability mass assigned to the decision region:
\begin{equation}
\rho_{\mathrm{dec}}
=\Pr_s(\theta \geq \theta_{\mathrm{dec}})
=\frac{s(1-\theta_{\mathrm{dec}})}
{\theta_{\mathrm{dec}}+s(1-\theta_{\mathrm{dec}})}.
\end{equation}
Experiments show that setting the decision-region probability mass to 50\% yields the best performance, which corresponds to $s=9.0$.

\section{Experiments}

\subsection{Dataset and Metrics}

We evaluate 4D-WAM on two challenging splits from NAVSIM benchmarks: \textit{navtest} and \textit{navhard}.
Both contain diverse and complex driving scenarios, while \textit{navhard} places greater emphasis on difficult cases and adopts a two-stage closed-loop evaluation protocol.
We adopt two official evaluation metrics: PDMS, proposed in NAVSIM v1, and EPDMS, proposed in NAVSIM v2. PDMS combines no at-fault collisions (NC), drivable-area compliance (DAC), ego progress (EP), time-to-collision (TTC), and comfort (C).
EPDMS extends PDMS with driving-direction compliance (DDC), traffic-light compliance (TLC), lane keeping (LK), and extended comfort (EC), while replacing C with history comfort (HC).


\subsection{Implementation Details}
Following the NAVSIM convention, we use three historical frames together
with the current observation as visual conditioning. Both the future video
and ego trajectory are predicted at $2$ Hz over a $4$-second horizon. For
image preprocessing, we crop $28$ pixels from the top and bottom of each
frame and resize it to $352\times640$. The entire network is trained under a flow-matching framework using two
stages. Stage I optimizes the joint video-action objective on
\textit{navtrain} for $40$ epochs. Stage II initializes from the Stage-I
checkpoint and continues training for another $40$ epochs with
4D consistency supervision. We use AdamW with a cosine learning-rate
schedule, with learning rates of $1\times10^{-4}$ and $2\times10^{-5}$
for Stages I and II, respectively. All models are trained on $16$ NVIDIA
H200 GPUs. Unless otherwise specified, inference uses $10$ denoising steps.

\subsection{Results}

\noindent\textbf{Results on \textit{navtest}.}
Table~\ref{tab:navsim_v2_epdms} presents the results under the NAVSIM v2
protocol. Using only single-view camera input, 4D-WAM achieves the highest
EPDMS of 90.6, outperforming the second-best method, DriveFine, by 0.9 points.
It also obtains the best NC, DDC, EP, and TTC scores, together with the
second-best DAC. These results demonstrate that 4D-WAM achieves a strong
overall balance among safety, driving compliance, and ego progress.

Table~\ref{tab:navsim_v1_pdms} reports the results under the NAVSIM v1
protocol. 4D-WAM achieves the highest PDMS of 90.9, surpassing ReCogDrive
and DriveVLA-W0 by 0.1 and 0.7 points, respectively. Notably, 4D-WAM uses supervised fine-tuning alone yet outperforms ReCogDrive, which employs reinforcement learning with a reward explicitly designed based
on the PDMS scoring rule.

\noindent\textbf{Results on \textit{navhard}.}
As shown in Table~\ref{tab:navsim_v2_navhard}, 4D-WAM achieves the highest
EPDMS of 35.9, surpassing the previous best result of 32.2 by 3.7 points.
In the real-observation stage, it achieves the best DDC, EP, and EC, while
in the synthetic follow-up stage, it ranks first in NC, DDC, EP, TTC, and
LK. These broad Stage-2 gains underscore the advantage of 4D consistent
world modeling, as coherent future geometry and motion support safer, more
compliant, and more progress-efficient planning from synthetic follow-up
observations.

\subsection{Ablation Studies}

\begin{figure*}[!t]
\centering
\includegraphics[width=\textwidth]{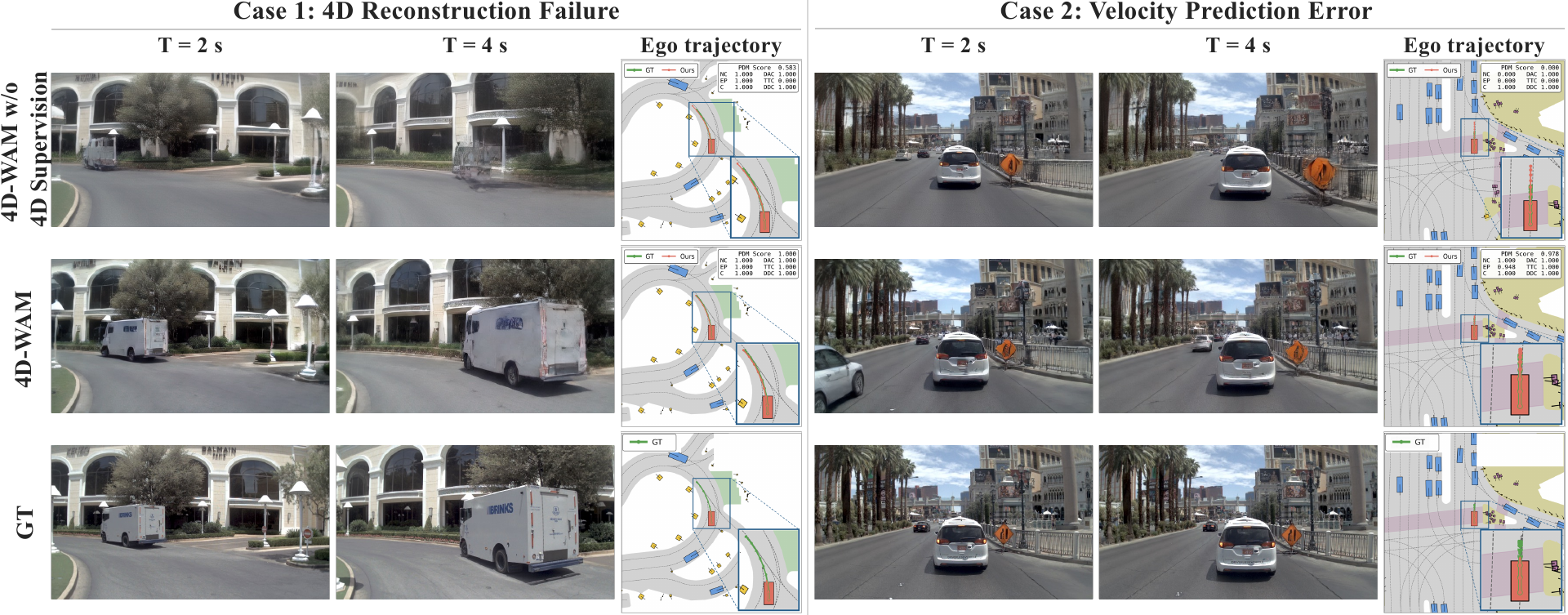}
\caption{\textbf{Effect of 4D consistency supervision.}
We compare 4D-WAM without 4D supervision (top), the full model (middle), and
ground truth (bottom). 4D consistency supervision improves the model's understanding of 4D scenes, enabling more coherent predictions of dynamic vehicles. It also enhances distance perception ability and avoids collision.}
\label{fig4}
\end{figure*}

\noindent\textbf{Qualitative Effects of 4D Consistency Supervision.}
Fig.~\ref{fig4} compares 4D-WAM with its ablated variant trained without
4D consistency supervision, denoted as \textit{4D-WAM w/o 4D Supervision}.
Case 1 presents a reconstruction failure involving a highly dynamic object.
Without 4D consistency supervision, the model fails to preserve the truck's
shape, scale, and displacement across time, resulting in inconsistent scene reconstruction and a deviated ego trajectory. With the proposed supervision,
4D-WAM reconstructs its geometry and motion more coherently, leading to a
more reasonable trajectory prediction. Case 2 illustrates a velocity prediction error. The model without
4D consistency supervision misestimates the velocity of the lead vehicle,
resulting in an overly aggressive driving strategy with a potential collision risk. In contrast, 4D consistency supervision enhances 4D-WAM's distance
perception, leading to more accurate velocity estimation and a safer ego
trajectory. Together,
these cases demonstrate that 4D consistency supervision improves both dynamic
scene reconstruction and distance-aware motion prediction, thereby benefiting trajectory planning.


\begin{table}[t]
\centering

\footnotesize
\setlength{\tabcolsep}{2.5pt}
\renewcommand{\arraystretch}{1.08}

\begin{tabular}{@{}lcccccc@{}}
\toprule
\textbf{Configuration}
& \textbf{NC} $\uparrow$
& \textbf{DAC} $\uparrow$
& \textbf{EP} $\uparrow$
& \textbf{LK} $\uparrow$
& \textbf{EC} $\uparrow$
& \textbf{EPDMS} $\uparrow$ \\
\midrule

WAM Base
& 98.3
& 97.0
& \underline{98.3}
& 86.4
& 86.4
& 88.8 \\

+ History
& 98.4
& 97.7
& \textbf{98.4}
& 86.7
& \underline{87.4}
& 89.6 \\

+ Feature Loss
& \underline{99.0}
& 97.6
& 98.1
& 86.8
& 87.0
& 90.1 \\

+ Depth Loss
& \underline{99.0}
& \underline{97.8}
& \underline{98.3}
& \textbf{87.0}
& 87.3
& \underline{90.4} \\

+ Sampling
& \textbf{99.1}
& \textbf{97.9}
& \underline{98.3}
& \underline{86.9}
& \textbf{87.5}
& \textbf{90.6} \\

\bottomrule
\end{tabular}

\caption{Component-wise ablation on \textit{navtest} under the
NAVSIM v2 protocol. Components are added cumulatively from top to
bottom, where ``Sampling'' denotes our Decision-Oriented
Timestep Sampling strategy.}
\label{tab:ablation_components}
\end{table}

\noindent\textbf{Ablation Study of Key Components.}
Table~\ref{tab:ablation_components} evaluates the contribution of each
component under the NAVSIM v2 protocol. Historical conditioning first improves
EPDMS from 88.8 to 89.6. The feature consistency loss provides a further
substantial gain to 90.1 and increases NC from 98.4 to 99.0, indicating that
constraining the high-level layout of the underlying 4D scene provides effective
structural guidance for trajectory planning. Adding the depth consistency loss
further improves EPDMS to 90.4, together with gains in DAC, EP, and LK,
demonstrating the benefit of preserving fine-grained geometric relationships
and relative depth ordering. Finally, Decision-Oriented Timestep Sampling
raises EPDMS to 90.6 and achieves the best NC, DAC, and EC by allowing 4D
supervision to more frequently shape joint prediction during driving-decision
formation.

\begin{table}[t]
\centering
\footnotesize
\setlength{\tabcolsep}{3pt}
\renewcommand{\arraystretch}{1.05}

\begin{tabular}{lcccccc}
\toprule
{\boldmath$\rho_{\mathrm{dec}}$}
& \textbf{10\%}
& \textbf{20\%}
& \textbf{30\%}
& \textbf{40\%}
& \textbf{50\%}
& \textbf{60\%} \\
\midrule
$s$
& 1.00 & 2.25 & 3.86 & 6.00 & 9.00 & 13.50 \\
EPDMS $\uparrow$
& 90.2 & 90.3 & 90.3 & \underline{90.5} & \textbf{90.6} & 90.1 \\
\bottomrule
\end{tabular}

\caption{Ablation of timestep sampling distribution}
\label{tab:timestep_allocation}
\end{table}

\noindent\textbf{Ablation Study of Timestep Sampling Distribution.}
We vary the probability mass $\rho_{\mathrm{dec}}$ assigned to the decision
region. As shown in Table~\ref{tab:timestep_allocation}, uniform sampling ($s=1$) assigns 10\% of
the samples to this region and achieves an EPDMS of 90.2. Increasing
$\rho_{\mathrm{dec}}$ generally improves performance, with EPDMS peaking at
90.6 when 50\% of the samples are allocated to the decision region. This
improvement demonstrates that more frequent sampling of high-noise timesteps
enables 4D consistency supervision to more directly participate in and
optimize driving-decision formation. However, further increasing the
allocation to 60\% lowers EPDMS to 90.1, suggesting that excessive emphasis on
the decision region compromises the learning of subsequent refinement.

\begin{table}[t]
\centering
\footnotesize
\setlength{\tabcolsep}{3pt}
\renewcommand{\arraystretch}{1.05}

\begin{tabular}{@{}lcccccc@{}}
\toprule
\textbf{Steps}
& \textbf{1}
& \textbf{2}
& \textbf{4}
& \textbf{6}
& \textbf{8}
& \textbf{10} \\
\midrule
EPDMS $\uparrow$
& 88.1 & 90.5 & 90.5 & 90.5 & 90.5 & 90.6 \\
Latency (s) $\downarrow$
& 0.22 & 0.31 & 0.44 & 0.55 & 0.67 & 0.79 \\
\bottomrule
\end{tabular}

\caption{Effect of denoising steps on planning performance and inference latency, evaluated on an NVIDIA RTX 4090.}
\label{tab:denoising_steps}
\end{table}

\noindent\textbf{Ablation Study of Denoising Steps.}
As shown in Table~\ref{tab:denoising_steps}, EPDMS rises sharply from
88.1 to 90.5 within two denoising steps and remains nearly saturated
thereafter, reaching 90.6 at ten steps. We attribute this rapid convergence
to the early-decision phenomenon, whereby the core driving decision is
established during the initial high-noise steps. Leveraging this property,
we reveal that 4D-WAM achieves planning performance comparable to ten-step
inference with only two denoising steps, while reducing latency from
0.79\,s to 0.31\,s.

\section{Conclusion}
In this work, we introduced 4D-WAM, a 4D consistent world-action model.
During training, 4D consistency supervision guides the model to capture and
predict geometrically and temporally coherent 4D scene dynamics, without
incurring additional inference cost. Motivated by the early-decision
phenomenon uncovered in WAMs, we further proposed Decision-Oriented Timestep
Sampling, enabling 4D consistency supervision to more effectively guide
driving-decision formation. 4D-WAM achieves state-of-the-art performance on
both NAVSIM v1 and v2. Qualitative results further demonstrate improved
dynamic object reconstruction and velocity prediction, leading
to safer and more accurate ego trajectories in complex driving scenarios.

\clearpage
\setlength{\bibsep}{0pt}
\bibliography{aaai2027}

\normalfont\normalsize
\setcounter{secnumdepth}{1}

\setcounter{section}{0}
\setcounter{subsection}{0}
\setcounter{figure}{0}
\setcounter{table}{0}
\setcounter{equation}{0}

\renewcommand{\thesection}{S\arabic{section}}
\renewcommand{\thesubsection}{S\arabic{section}.\arabic{subsection}}
\renewcommand{\thefigure}{S\arabic{figure}}
\renewcommand{\thetable}{S\arabic{table}}
\renewcommand{\theequation}{S\arabic{equation}}

\clearpage
\twocolumn[
\begin{center}
{\LARGE\bfseries
4D-WAM: 4D Consistent World Modeling for Autonomous Driving
\par}
\vspace{0.5em}
{\Large\bfseries Supplementary Material\par}
\end{center}
\vspace{0.8em}
]


\begin{figure*}[!t]
    \centering
    \includegraphics[
        width=\textwidth,
        height=0.9\textheight,
        keepaspectratio
    ]{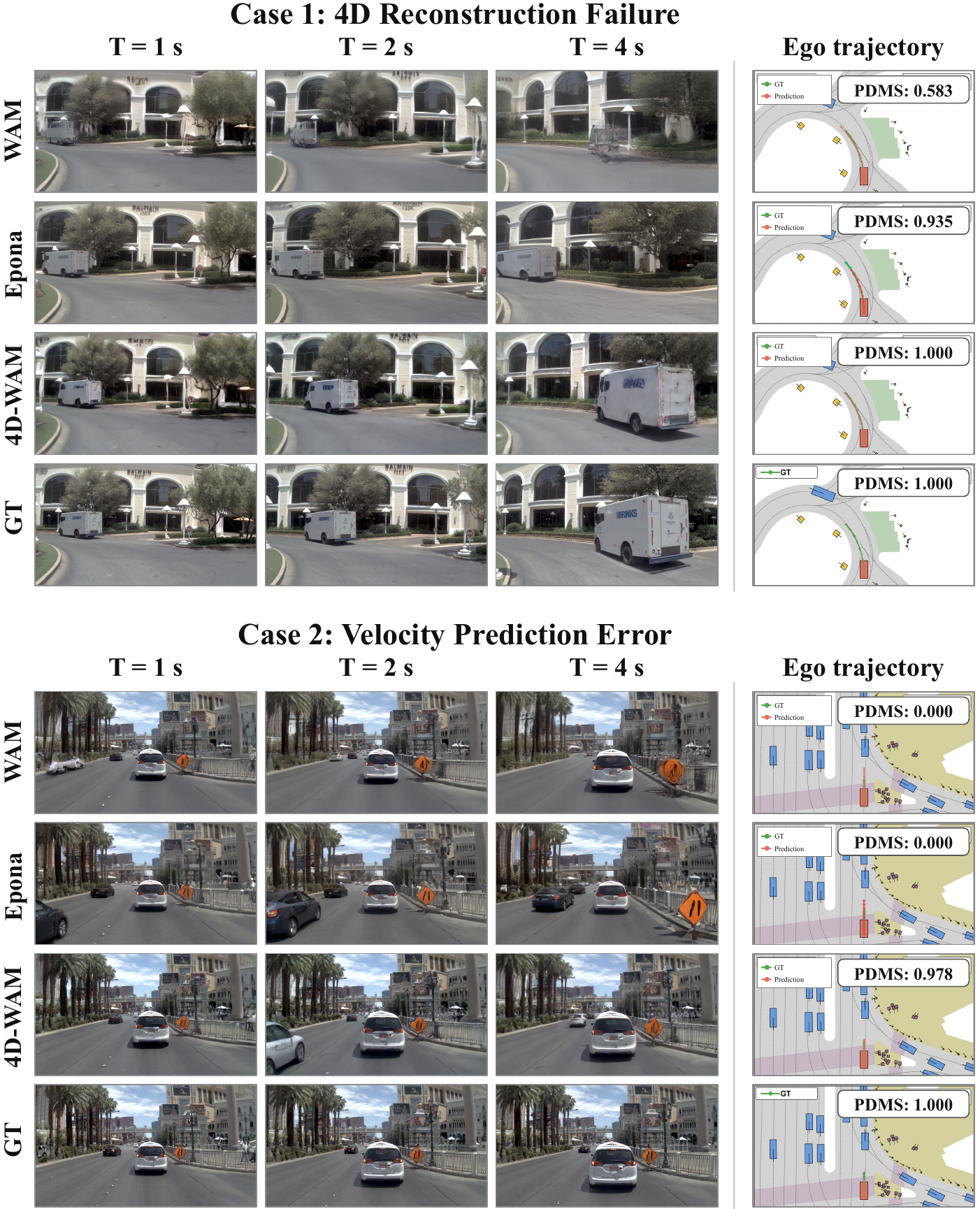}
    \caption{
        Qualitative comparison among WAM, Epona, 4D-WAM, and the ground truth
        on two representative NAVSIM cases. WAM denotes the 4D-WAM backbone trained without 4D consistency supervision.
        Epona is evaluated using its default 100-step denoising setting, and WAM and 4D-WAM use the 10-step denoising configuration adopted in the main paper.
    }
    \label{fig:epona_qualitative_comparison}
\end{figure*}

This supplementary material provides additional implementation details and
qualitative results to support the main paper:
\begin{itemize}
    \item Sec.~\ref{sec:supp_4d_supervision} provides implementation details
    of the feature- and depth-consistency losses used for 4D consistency supervision.
    \item Sec.~\ref{sec:supp_epona} presents additional qualitative comparisons
    with Epona on representative NAVSIM scenarios.
    \item Sec.~\ref{sec:zero_shot_generalization} evaluates the zero-shot
    generalization capability of 4D-WAM on Waymo.
\end{itemize}

\section{Implementation Details of 4D Consistency Supervision}
\label{sec:supp_4d_supervision}

We provide additional implementation details for the feature and depth
consistency losses introduced in the main paper.


\subsection{Feature Consistency Loss}

As introduced in the main paper, the feature consistency loss is formulated as
\begin{equation}
\mathcal{L}_{\mathrm{feat}}
=
\frac{1}{|\mathcal{S}|}
\sum_{\ell\in\mathcal{S}}
d_{\cos}
\left(
\hat{\mathbf{F}}^{\ell},
\mathbf{F}_{\mathrm{gt}}^{\ell}
\right),
\end{equation}
where we select $\mathcal{S}=\{11,17,23\}$ to extract representations at
different depths of VGGT-$\Omega$. In implementation, the representation at
each selected layer is further divided into three token groups:
\begin{equation}
\mathbf{F}^{\ell}
=
\left[
\mathbf{F}_{\mathrm{cam}}^{\ell};
\mathbf{F}_{\mathrm{reg}}^{\ell};
\mathbf{F}_{\mathrm{patch}}^{\ell}
\right].
\end{equation}
This partition follows the native VGGT-$\Omega$ \cite{vggt_omega} architecture, whose camera,
register, and patch tokens encode complementary camera, global-context, and
local-geometry information, respectively. We compute the cosine distance independently for each group,
which gives the following expanded implementation of
$\mathcal{L}_{\mathrm{feat}}$:
\begin{equation}
\mathcal{L}_{\mathrm{feat}}
=
\frac{1}{|\mathcal{S}|}
\sum_{\ell\in\mathcal{S}}
\sum_{g\in\mathcal{T}}
d_{\cos}
\left(
\hat{\mathbf{F}}_{g}^{\ell},
\mathbf{F}_{\mathrm{gt},g}^{\ell}
\right),
\end{equation}
where
$\mathcal{T}=\{\mathrm{cam},\mathrm{reg},\mathrm{patch}\}$, and each
distance is averaged over the future-frame and token dimensions. All selected
layers and token groups are assigned equal weights.

For the patch-token group, the $352\times640$ input resolution and patch size
of $16$ produce a $22\times40$ spatial token grid. Before computing the cosine
distance, we apply $2\times2$ average pooling to obtain an $11\times20$ grid.
This reduces the memory cost of geometric supervision while retaining the
spatial organization needed to constrain local object geometry. Consequently,
the three feature groups jointly supervise view-level scene configuration,
global multi-frame structure, and local geometric details.

\subsection{Depth Consistency Loss}

As described in the main paper, the depth consistency loss is evaluated over
a reliable future-pixel set $\mathcal{Q}$. For each retained pixel, we first
compute the predicted and target log-depths:
\begin{equation}
\hat d_q=\log(\hat D_q+\epsilon),
\qquad
d_{\mathrm{gt},q}=\log(D_{\mathrm{gt},q}+\epsilon).
\end{equation}
The depth consistency loss is then written as
\begin{equation}
\mathcal{L}_{\mathrm{depth}}
=
\frac{1}{|\mathcal{Q}|}
\sum_{q\in\mathcal{Q}}
\mathrm{SmoothL1}_{\beta}
\left(
\hat d_q-d_{\mathrm{gt},q}
\right).
\end{equation}

We construct $\mathcal{Q}$ independently for each sample $i$ and future frame
$t$. We first collect an initial valid set $\mathcal{B}_{i,t}$ containing
pixels for which the predicted depth, target depth, and VGGT-$\Omega$
confidence are finite, both depth values are positive, and the target
confidence is greater than $10^{-5}$.

\begin{figure*}[t]
    \centering
    \includegraphics[width=\textwidth]{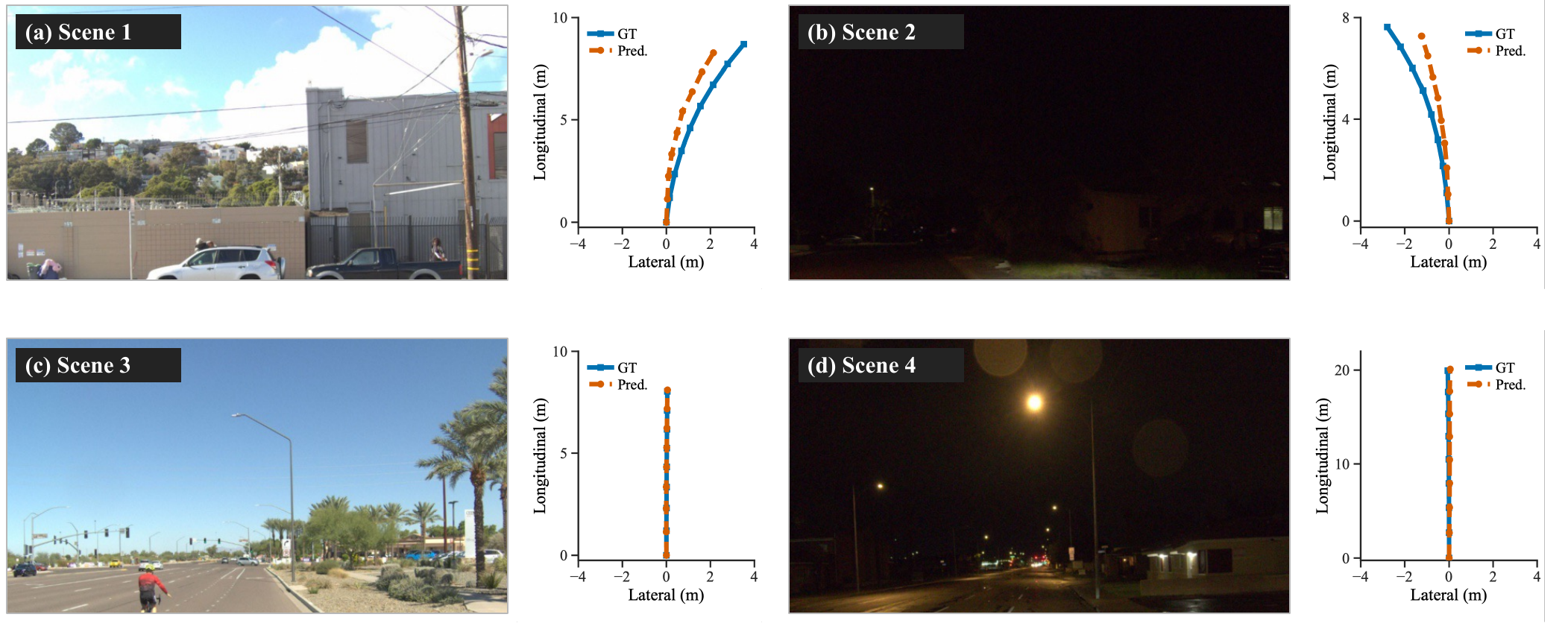}
    \caption{\textbf{Zero-shot generalization on Waymo.}
    Without training or fine-tuning on Waymo, 4D-WAM generates ego trajectories
    that closely align with the ground truth across diverse maneuvers and
    illumination conditions, including turning, straight driving, and
    challenging nighttime scenes.}
    \label{fig:waymo_zeroshot}
\end{figure*}

We then determine a frame-adaptive confidence threshold by filtering out the
lowest-confidence $30\%$ of the valid pixels. Let $Q_{\alpha}$ denote the
empirical $\alpha$-quantile. The confidence threshold is defined as
\begin{equation}
c_{i,t}
=
Q_{0.3}
\left(
\left\{
C_{\mathrm{gt},i,t,q}
\mid q\in\mathcal{B}_{i,t}
\right\}
\right),
\end{equation}
and the confidence-filtered set is
\begin{equation}
\mathcal{C}_{i,t}
=
\left\{
q\in\mathcal{B}_{i,t}
\mid
C_{\mathrm{gt},i,t,q}\geq c_{i,t}
\right\}.
\end{equation}
We use $\mathcal{Q}_{i,t}=\mathcal{C}_{i,t}$ as the reliable-pixel set for
the corresponding frame. The complete set $\mathcal{Q}$ is formed by
collecting the retained pixels over all samples and future frames. Computing
the confidence threshold separately for each frame allows the filtering
process to adapt to scene-dependent variations in confidence. We use
$\beta=0.1$ for the Smooth-L1 loss and $\epsilon=10^{-6}$ for the logarithmic
depth transformation.

We set $\lambda_{\mathrm{feat}}=\lambda_{\mathrm{depth}}=1$. During the
first $0.5$ epoch of Stage II, the resulting 4D consistency loss is gradually
introduced using a half-cosine warm-up:
\begin{equation}
w_{\mathrm{warm}}(r)
=
\frac{1-\cos(\pi r)}{2},
\qquad
r\in[0,1],
\end{equation}
where $r$ denotes the normalized progress within the warm-up period. The
weight remains one thereafter.

\section{Additional Qualitative Comparison}
\label{sec:supp_epona}

We complement the qualitative analysis in the main paper with Epona~\cite{epona}, another world-action model that jointly predicts future videos and ego trajectories. All methods use the same four observed frames, including three history frames and the current observation. Following their respective default settings, Epona uses 100 denoising steps, whereas WAM and 4D-WAM use the 10-step denoising configuration adopted in the main paper. Since the released Epona model predicts trajectories over only a 3-second horizon, we adapt it to the 4-second NAVSIM horizon by performing two consecutive trajectory predictions, retaining the first 2 seconds from each prediction, and concatenating the resulting segments. We compare temporally aligned video predictions at $T\in\{1,2,4\}\,\mathrm{s}$.

\paragraph{Case 1: 4D Reconstruction Failure.}
As shown in Fig.~\ref{fig:epona_qualitative_comparison}, WAM completely fails to reconstruct the fast-moving truck, producing severe visual artifacts as its structure collapses over time. Epona preserves a recognizable appearance but fails to maintain the truck's rigid geometry, with its body becoming increasingly stretched over the prediction horizon. It also fails to correctly extrapolate the truck's motion from the historical observations. This inaccurate motion prediction leads to an overly conservative ego trajectory, resulting in insufficient ego progress and unnecessarily slow driving. In contrast, 4D-WAM predicts the truck's shape and motion more coherently and produces an ego trajectory closely aligned with the ground truth.

\paragraph{Case 2: Velocity Prediction Error.}
As shown in Fig.~\ref{fig:epona_qualitative_comparison}, both WAM and Epona produce visually plausible future frames but incorrectly predict the velocity of the lead vehicle. This error results in overly aggressive driving strategies with a potential collision risk, leading to a PDMS of 0.000 for both methods. In contrast, 4D-WAM models the lead vehicle's motion more accurately and produces a safer ego trajectory, achieving a PDMS of 0.978.

\section{Zero-Shot Generalization Capability}
\label{sec:zero_shot_generalization}

To examine whether 4D-WAM learns transferable driving knowledge beyond its
training distribution, we conduct a qualitative zero-shot evaluation on Waymo \cite{waymo}
validation sequences. The evaluated checkpoint is trained exclusively on
NAVSIM, without using any Waymo data for training or fine-tuning.

\paragraph{Cross-domain trajectory planning.}
As shown in Fig.~\ref{fig:waymo_zeroshot}, despite differences in
data-collection hardware, camera configurations, and image resolutions between
NAVSIM and Waymo, 4D-WAM produces accurate ego trajectories across diverse
driving scenarios, including both straight driving and turning maneuvers.
These results demonstrate strong zero-shot transfer from NAVSIM to Waymo
despite substantial appearance and road-layout shifts.

\paragraph{Robustness in challenging scenes.}
Scenes~2 and~4 evaluate challenging nighttime conditions with severely limited
visibility, which are scarcely represented in NAVSIM. Despite the substantial
illumination shift, 4D-WAM preserves the correct driving intent and
longitudinal progress. This robustness suggests that 4D consistency supervision
promotes transferable geometric and temporal reasoning rather than reliance on
dataset-specific appearance cues.



\end{document}